\documentclass[10pt,twocolumn,letterpaper]{article}

\usepackage{cvpr} 
\usepackage{times}
\usepackage{epsfig}
\usepackage{graphicx}
\usepackage{amsmath}
\usepackage{amssymb}
\usepackage[pagebackref]{hyperref}
\usepackage{booktabs}
\hypersetup{
    colorlinks=true,
    linkcolor=blue,
    citecolor=blue,
    urlcolor=blue
}
\usepackage{float}
\usepackage{color}  
\usepackage{multirow}
\usepackage{threeparttable}

\graphicspath{{imgs/}}
\usepackage{cite}
\newcommand{\tabincell}[2]{\begin{tabular}{@{}#1@{}}#2\end{tabular}}

\begin{document}

\title{Improving Contactless Fingerprint Recognition \\ with Robust 3D Feature Extraction and  Graph Embedding}

\author{    Yuwei Jia \quad
    Siyang Zheng \quad
    Fei Feng \quad
    Zhe Cui\textsuperscript{*} \quad
    Fei Su \\
    Beijing Key Laboratory of Network System and Network Culture \\Beijing University of Posts and Telecommunications, Beijing, China \\
    {\tt\small \{jiayw, zhengsiyang, ffei, cuizhe, sufei\}@bupt.edu.cn}
  \thanks{Zhe Cui is the corresponding author. This work is supported in part by the National Natural Science Foundation of China under Grants 62206026.}
}

\maketitle
\thispagestyle{empty}

\begin{abstract}
    Contactless fingerprint has gained lots of attention in recent fingerprint studies. However, most existing contactless fingerprint algorithms treat contactless fingerprints as 2D plain fingerprints, and still utilize traditional contact-based 2D fingerprints recognition methods. This recognition approach lacks consideration of the modality difference between contactless and contact fingerprints, especially the intrinsic 3D features in contactless fingerprints. This paper proposes a novel contactless fingerprint recognition algorithm that captures the revealed 3D feature of contactless fingerprints rather than the plain 2D feature. The proposed method first recovers 3D features from the input contactless fingerprint, including the 3D shape model and 3D fingerprint feature (minutiae, orientation, etc.). Then, a novel 3D graph matching method is proposed according to the extracted 3D feature. 
    Additionally, the proposed method is able to perform robust 3D feature extractions on various contactless fingerprints across multiple finger poses.
    The results of the experiments on contactless fingerprint databases show that the proposed method successfully improves the matching accuracy of contactless fingerprints. Exceptionally, our method performs stably across multiple poses of contactless fingerprints due to 3D embeddings, which is a great advantage compared to 2D-based previous contactless fingerprint recognition algorithms.
\end{abstract}

\section{Introduction}

Compared with traditional contact-based fingerprint, which relies on pressing finger skin against sensors to gather fingerprint images, contactless fingerprint provides a touchless approach to acquiring fingerprint images. Since the collection process is touch-free, the gathered fingerprints are of no skin distortion, which is a crucial problem in contact-based fingerprint matching \cite{bazen2003matching}. The collection process is more hygienic and convenient and can even be done with mobile phone cameras \cite{stein2012fingerphoto}\cite{sankaran2015smartphone}.

Despite these advantages, contactless fingerprint recognition still meets challenges on modality differences with contact-based fingerprints \cite{tan2021minu}. The contactless fingerprint image is acquired by a camera-like imaging system \cite{lee2006study}\cite{parziale2006surround} in 3D space, which differs from traditional contact-based 2D sensors. The finger surface curvature is incorporated into the imaging result, which leads to evident perspective distortion on contactless fingerprints \cite{labati2013contactless}\cite{tan2020towards}. The distortion causes changes in orientation, period, and minutiae location, thus degrading recognition performances \cite{jain1997line}.  

\begin{figure}[!t]
\begin{center}
\centerline{\includegraphics[width=\linewidth]{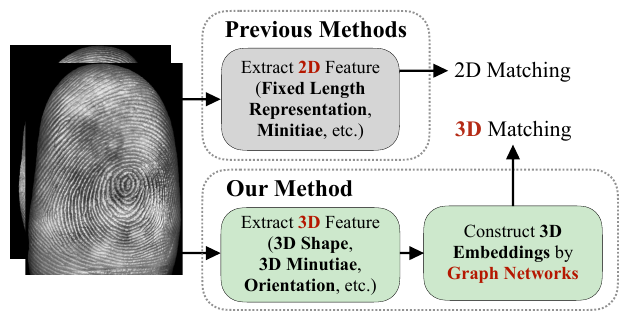}}
\vspace{-20pt} 

\end{center}

\caption{Compared with previous methods that only uses 2D fingerprint feature, our method extracts 3D features of contactless fingerprints and encodes them via a 3D graph neural network for fingerprint matching.}
\label{fig:intro}
\vspace{-10pt} 
\end{figure}

To overcome perspective distortion and accurately match contactless fingerprints, researchers have proposed methods to deal with this issue. A significant approach is to treat perspective distortion similar to elastic skin distortion \cite{si2015detection} in contact-based 2D fingerprints. A transformation network \cite{lin2018matching}\cite{Dabouei2019}\cite{grosz2022contact}\cite{dong2023syn} is trained to fit the distortion, thus rectifying the fingerprint to the undistorted level. This approach is generally used in contactless-to-contact (CL2C) matching, since it treats a contactless fingerprint as a distorted 2D fingerprint, similar to a contact-based 2D fingerprint. 

As for contactless-to-contactless (CL2CL) matching, several studies have further considered the intrinsic 3D information of contactless fingerprints. A finger shape model \cite{tan2020towards}\cite{tan2021minu}\cite{cui2023monocular} is recovered from the original contactless fingerprint. Then, all the contactless fingerprints are rotated to the same pose (usually the front pose) in 3D space to eliminate pose variations for improving the matching accuracy.

Although significant progress has been made in contactless fingerprint recognition, the modality characteristics of contactless fingerprints have yet to be fully studied. Most algorithms \cite{lin2018matching}\cite{Dabouei2019}\cite{lin2019cnn}\cite{grosz2022contact}\cite{dong2023syn} treat contactless fingerprints as plain 2D fingerprint images, therefore neglecting the 3D information hidden in contactless fingerprints. Some studies further consider 3D information. However, the recovered 3D shape is only used for pose correction \cite{labati2013contactless}\cite{tan2020towards} or unwarping \cite{Sollinger2021}\cite{cui2023monocular}. The rotated or unwarped contactless fingerprint is still treated as a 2D image for the following feature extraction and matching steps. 

In this paper, we propose a novel contactless fingerprint recognition method that uses 3D information in contactless fingerprints. As shown in Fig. \ref{fig:intro}, unlike previous methods that only extracted 2D features from contactless fingerprint images, the proposed method takes a contactless fingerprint as input, and then extracts its 3D feature, including the shape model, 3D minutiae feature, etc. Based on those features, we can reconstruct the feature points in 3D space and extract 3D feature embeddings. These 3D features are finally used for 3D fingerprint graph matching. 

In general, our work has four contributions:
\vspace{-5pt} 

\begin{itemize}

\item[1)] This proposed method innovatively extracts features for contactless fingerprint recognition within a 3D space. The proposed method inputs a single 2D fingerprint image, and is able to output 3D features and 3D matching results. In contrast to previous contactless fingerprint methods \cite{tan2020towards}\cite{tan2021minu}\cite{grosz2022contact}\cite{cui2023monocular}\cite{dong2023syn}\cite{shi2022towards} that only use 2D features during matching, our method maintains 3D features throughout the algorithm.
\vspace{-5pt} 
\item[2)] Our method improves the monocular depth estimation network \cite{cui2023monocular}, enhances the depth prediction accuracy of previous methods, and addresses the issue of poor generalization of previous methods to various contactless fingerprint datasets.
\vspace{-5pt} 
\item[3)] Our work makes use of the commonly used graph neural network \cite{li2019deepgcns}\cite{Wu2021Survey} that can represent the complicated relationships within structured data and has been successfully applied in fingerprint matching \cite{su2023gnn}. Our work further explores its application in contactless fingerprint and 3D fingerprint recognition.
\vspace{-5pt} 
\item[4)] We conduct thorough experiments on contactless fingerprint datasets with significant pose variations. Experiment results on several datasets prove that our method can handle the pose variation and perspective distortion issues in contactless fingerprint recognition, and reaches the SOTA matching performances.

\end{itemize}

\begin{figure*}[!t]
\begin{center}
\centerline{\includegraphics[width=0.9\linewidth]{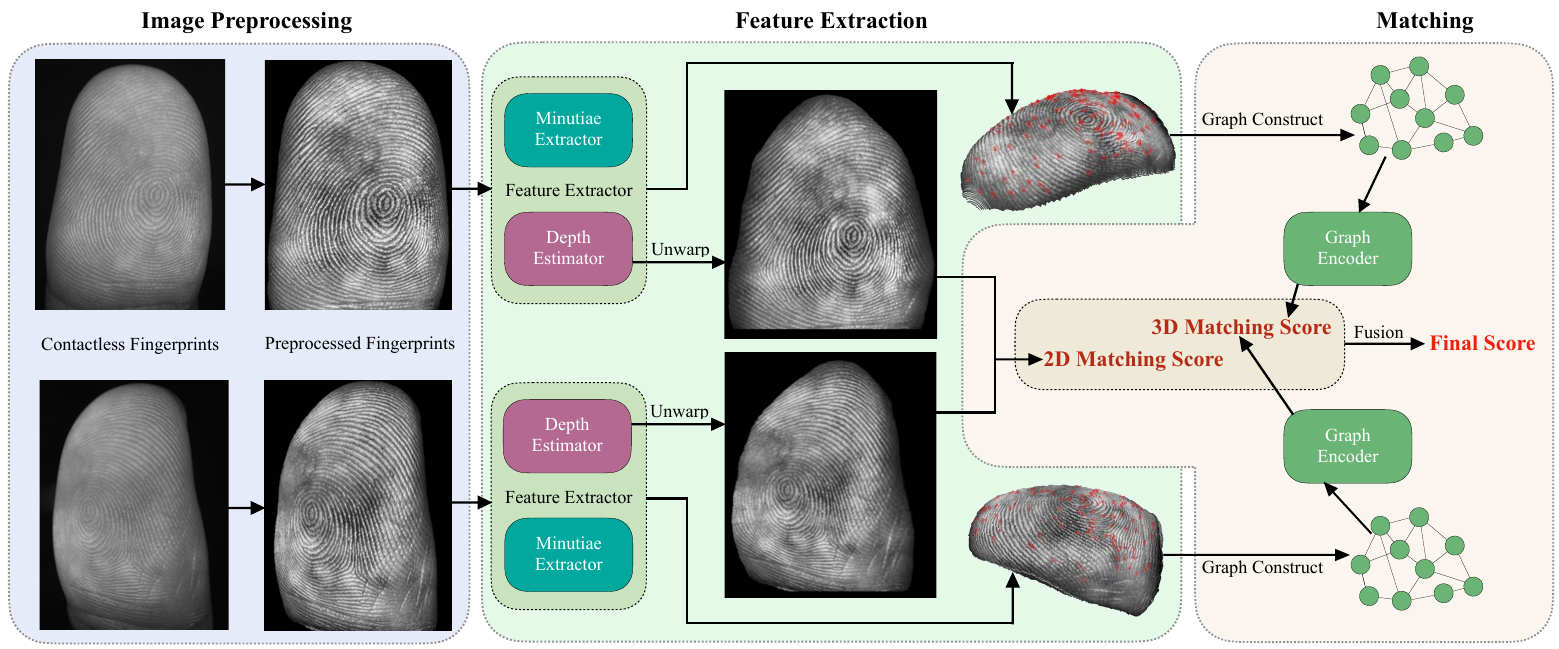}}
\vspace{-20pt} 

\end{center}

\caption{The pipeline of the proposed method. Our method first extracts 3D features (3D shape, 3D minutiae feature, deep features, etc.)
 from the input contactless 2D fingerprint, then uses the extracted 3D features to conduct matching both in 2D space with unwarping and in 3D space by extracting fixed length graph embedding.}
\label{fig:base_idea}
\vspace{-10pt} 
\end{figure*}
\section{Related Work}

\subsection{Contactless fingerprint recognition with only 2D feature}
Early studies \cite{song2004new}\cite{lee2006study}\cite{hiew2007touch}\cite{parziale2008touchless} dealt with contactless fingerprints the same as conventional contact-based fingerprints in feature extraction and matching methods. Further work \cite{chen2006touchless}\cite{zhou2013performance}\cite{zhou2014performance} noticed the the fingerprint quality and perspective distortion issues of contactless fingerprints and made several improvements.

For contactless fingerprint recognition, specific feature extraction and matching methods are designed with the deep learning technique. Algorithms including segmentation \cite{Dabouei2019}\cite{Malhotra2020}, enhancement \cite{lin2019cnn}\cite{Dabouei2019}\cite{grosz2022contact}, minutiae extraction \cite{lin2017multi}\cite{tan2020towards}\cite{tan2021minu}\cite{zhang2021multi}, and matching \cite{lin2018matching}\cite{lin2019cnn}\cite{grosz2022contact} are proposed to improve the recognition performance. Shi et al. \cite{shi2022towards} adopted graph neural networks for contactless fingerprint matching, just as we did. However, they did not construct a graph in 3D space. Moreover, their method ultimately matched the minutiae of contactless fingerprints, whereas our method encodes three-dimensional fingerprints into fixed-length features.

The perspective distortion is another major issue in contactless fingerprint recognition. Early research \cite{lee2006study}\cite{hiew2007touch}\cite{kumar2011contactless}\cite{labati2013contactless} only used the central region or small rotation image of contactless fingerprints to decrease the impact of perspective distortion. A more general method is to train a distortion correction model \cite{lin2018matching}\cite{Dabouei2019}\cite{grosz2022contact}\cite{dong2023syn} to stimulate the deformation grid, which is the same as in skin distortion rectification \cite{si2015detection}\cite{Dabouei2018} in contact-based 2D fingerprint.

In general, 2D-based contactless fingerprint recognition methods have made significant improvements, especially with the deep learning approach. But the 2D-based method fails to capture the 3D feature of contactless fingerprints, therefore has disadvantages in recognizing fingerprints of large 3D variations \cite{tan2020towards}. Also, learning the distortion field benefits the CL2C matching only, and is unsuitable for CL2CL matching.
\subsection{Contactless fingerprint recognition with 3D feature}
Recently, a few studies \cite{tan2020towards}\cite{Sollinger2021}\cite{cui2023monocular} have noticed the 3D information in contactless fingerprints and attempted to fuse 3D features into recognition. Usually, a finger-shaped model is estimated to stimulate the 3D finger. The reconstructed 3D shape is then used for pose correction \cite{stein2012fingerphoto}\cite{labati2013contactless}\cite{tan2020towards} or 3D-2D unwarping \cite{fatehpuria2006acquiring}\cite{chen2006touchless}\cite{Shafaei2009new}\cite{Sollinger2021}\cite{cui2023monocular}. However, the rotated or unwarped fingerprint is still a 2D image. The consequent feature extraction and matching algorithms are operated on the processed 2D contactless fingerprint. Therefore, the 3D information utilized here cannot be explicitly called '3D feature', as the fingerprint feature is yet 2D-based. The benefits of 3D information are not used in the final matching process.

Several 3D fingerprint recognition algorithms have developed specific 3D features in 3D fingerprint matching, such as 3D minutiae \cite{kumar2013towards}\cite{lin2017tetrahedron}, 3D ridge feature \cite{yin2021fingerprint}, and curvature \cite{liu2015study}\cite{galbally2017full}. However, the fingerprints used in these studies are 3D fingerprints acquired from specific 3D equipment like shape-from-shading \cite{kumar2013towards}\cite{lin2017tetrahedron} or shape-from-multiview \cite{liu2015study}, which are a bit different and more complicated than the contactless 2D fingerprints gathered by camera. Most existing image-based 3D fingerprint algorithms require two or more fingerprint captures and a calibrated camera system to restore the 3D information, which are too reluctant to be applied on the usual contactless fingerprint recognition scenes.



\section{Methods}
Fig. \ref{fig:base_idea} shows the framework of the proposed method. Our method consists of three main steps: 1) the preprocessing step that segments, enhances, and unifies the scale and pose angles of fingerprint images \cite{Dabouei2019}\cite{cui2023monocular}; 2) the feature extraction step that goes through a multitask network to output the 3D shape and minutiae features; 3) the matching step that uses the 3D feature and a 3D graph matching network \cite{su2023gnn} to output the matching scores. Although the input contactless fingerprint is 2D, our method can recover the 3D shape model, and therefore conduct a 3D feature matching algorithm. 

The score of 3D graph matching offers a new perspective for 2D fingerprint matching and can effectively enhance the performance of contactless fingerprint recognition. Meanwhile, we use our newly proposed monocular depth estimation network with better generalization to unwarp contactless fingerprints, correcting perspective distortion and matching them in a 2D space. By combining the two matching approaches, our method can effectively boost the performance of contactless fingerprint recognition.

\subsection{Preprocessing}\label{ssec:preprocess}
Due to the complicated situations of contactless fingerprints, it is essential to minimize the variations within training data to strengthen the robustness of the proposed network. Similar to previous studies \cite{Dabouei2019}\cite{grosz2022contact}\cite{cui2023monocular}\cite{dong2023syn}, we primarily standardize contactless fingerprint images from four perspectives: segmentation, contrast, pose and ridge frequency.

In fingerprint recognition tasks, attention is usually focused only on the first phalanx. Similar to Dong and Kumar \cite{dong2023syn}, we employ SOLOv2 \cite{wang2020solov2} for the segmentation of the first phalanx. We annotated the first 912 images in the ZJU dataset for SOLOv2 training and validation, and segmented the remaining part of the ZJU dataset, as well as the UWA \cite{zhou2014benchmark} and CFPose \cite{tan2020towards} datasets. We then employ the same method as in \cite{cui2023monocular} to adjust fingerprint contrast, orient the fingerprints correctly, and unify ridge frequency.


\begin{figure*}[!t]
\begin{center}
\centerline{\includegraphics[width=0.9\linewidth]{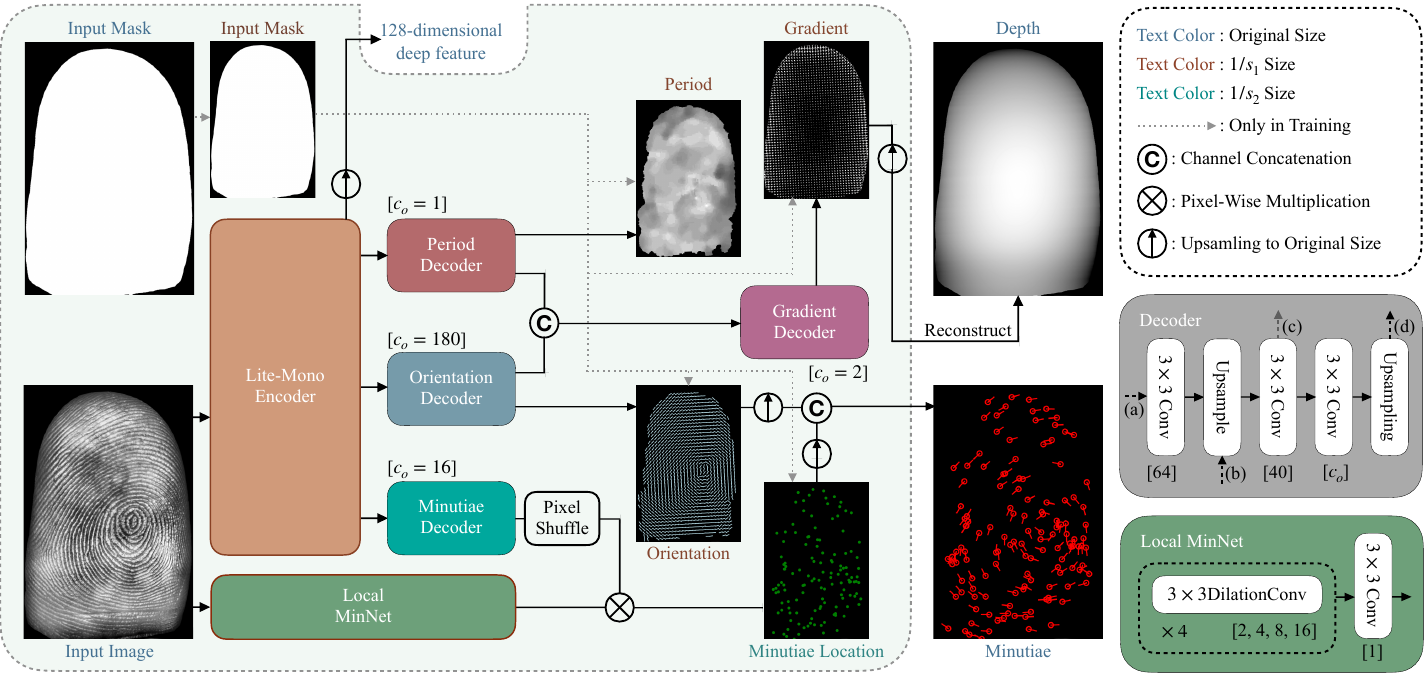}}
\vspace{-10pt} 

\end{center}
\caption{The detailed pipeline of the proposed 3D feature network, which extracts 3D features from the input contactless 2D fingerprint. The network outputs a 128-dimensional
deep feature along with the depth and minutiae feature for the following graph embedding. Each decoder (period, orientation, gradient, minutiae) contains 2 inputs and 1 output: (a) The deepest features of Lite-Mono Encoder as input. (b) The middle-level features of Lite - Mono Encoder that are used as input. (d) Final output. And (c) is the intermediate output only in period and orientation decoder. }
\label{fig:network1}
\vspace{-10pt} 

\end{figure*}

\subsection{Fingerprint Feature Extraction}
The 3D feature network takes a preprocessed contactless fingerprint as input, and then outputs the surface gradient, minutiae location, ridge period and orientation. It should be noted that the direct results from the network are in 2D. The estimated surface gradient is used for 3D reconstruction \cite{cui2023monocular} to get a 3D finger shape. Then, the minutiae and orientation are recomputed according to the reconstructed 3D shape, and we can finally get the 3D minutiae result.

Fig. \ref{fig:network1} shows the pipeline of the proposed feature extraction network. The network inputs the preprocessed fingerprint and segmentation mask from the above preprocessing step, and outputs the estimated gradient, minutiae, ridge period, and orientation. This network is an adaptation from \cite{cui2023monocular} with an extra minutiae branch, which is inspired by FingerNet \cite{fingernet2017}. This network has multitask branches, which has been proven in \cite{cui2023monocular} that a multitask network performs better than a single task. 

However, the method in \cite{cui2023monocular} still cannot perform depth estimation well on various contactless fingerprint datasets under different acquisition conditions, and the resulting depths are often not smooth enough. Therefore, we improved this network by adopting the Lite-Mono encoder from \cite{zhang2023lite} and fine-tuning it with pre-trained parameters, which enhanced the generalization ability of the previous network. Compared to standard Litemono encoder, our Litemono encoder only outputs feature maps of two scales, with 80 and 128 channels respectively. For each decoder, the 128-channel feature map serves as the input to the first convolutional layer, and the 80-channel feature map is concatenated with the output of the second convolutional layer to form the input to the third convolutional layer.

The four decoders (period, orientation, gradient, minutiae) process different output features, and are of the same structure with the only difference being the number of output channels. The structure of decoders is shown in the middle of the right side of Fig. \ref{fig:network1}, which is a series of convolution and upsampling layers. The input arrow to the second downsampling layer represents the skip connection of the Lite-Mono layer. The first output arrow represents the feature of the previous layer, and the feature of the direction field and period map is used for predicting the gradient. The second output arrow represents the final prediction result.

The minutiae branch uses the Local MinNet from the minutiae extraction network in \cite{feng2023detecting} for minutiae localization, as Lite-Mono Encoder \cite{zhang2023lite} is already capable of extracting global information from the input fingerprint effectively. Finally, the output minutiae are obtained by multiplying the local prediction results from the minutiae branch with the global prediction results from the depth estimation network. The minutiae map branch only outputs the minutiae location. The minutiae orientation is the same from the orientation branch, only expanding $180^{\circ}$ to $360^{\circ}$.



The network has four outputs: orientation, period, gradient, and minutiae. Therefore, the network loss also comprises four individual losses: orientation loss $\mathcal{L}_{\text{Ori}}$, period loss $\mathcal{L}_{\text{Ped}}$, gradient loss $\mathcal{L}_{\text{Grad}}$, and minutiae loss $\mathcal{L}_{\text{Minu}}$.

The orientation loss $\mathcal{L}_{\text{Ori}}$, period loss $\mathcal{L}_{\text{Ped}}$, and gradient loss $\mathcal{L}_{\text{Grad}}$ are the same as in \cite{cui2023monocular}, where detailed descriptions can be found. We construct the minutiae location map in the same manner as \cite{feng2023detecting} and adopt their proposed Dynamic End-to-End Loss Design as our $\mathcal{L}_{\text{Minu}}$.

Finally, the total loss is the sum of all:
\begin{equation}
\begin{aligned}
\mathcal{L}=\lambda_1\mathcal{L}_{\text{Ori}}+\lambda_2\mathcal{L}_{\text{Ped}}+\lambda_3\mathcal{L}_{\text{Grad}}+\lambda_4\mathcal{L}_{\text{Minu}}.
\end{aligned}
\label{equ:loss}
\end{equation}
$\lambda_1$-$\lambda_4$ are the weights of these four losses. All parameters used in the loss function are same as Cui et al. \cite{cui2023monocular} except $\lambda_4 = 1000$.


We upsample the output gradients, minutiae location maps, orientations, and 128-dimensional depth feature maps to the original image size. The gradients are used to reconstruct depth, while orientations and minutiae maps are transformed into 2D minutiae using the method from \cite{feng2023detecting}. Finally, gradients, depth, 2D minutiae, and depth feature maps are combined to construct the 3D fingerprint features for subsequent graph embedding.

\subsection{3D Feature Graph Embedding}

\textbf{3D Minutiae Feature.} \label{ssection:3d_feature}
The direct outputs of the feature extraction network are 2D. They need to be transformed into 3D space to fit for the 3D graph matching. 
Compared with 2D minutiae, 3D minutiae have an extra dimension, which needs to be calculated out.





The 3D depth $z$ is recovered from the gradient results $g_x$ and $g_y$ using numerical integration in \cite{cui2023monocular}, as the gradient is the differential of the surface depth. The calculated finger depth represents the $z$ coordinate, and is combined with the original image pixel $(x,y)$ to get a 3D coordinate $\mathbf{p}=(x,y,z)$. The direction of 3D minutiae can be obtained through a two-step rotation in the local coordinate system. The minutiae direction $\theta$ is the angle of the 3D minutiae in the x-y plane, which is the angle of the first step rotation. The gradients $g_x$ and $g_y$ represent the changes in axis $z$ respectively when $x$ and $y$ changes by one pixel. Therefore, the change $d_\theta$ in one pixel along the $\theta$ direction is 
\begin{equation}
    d_\theta = \cos\theta \frac{g_x}{\sqrt{g_x^2+1}} + \sin\theta \frac{g_y}{\sqrt{g_y^2+1}}
\label{equ:angle_phi}
\end{equation}
 This change can represent the tangent value of the angle $\varphi$ in the second step, that is, $\varphi=\arctan d_\theta$. Thus, based on the spherical coordinate system, the final direction of the 3D minutia is:
\begin{equation}    
\begin{aligned}
 \begin{bmatrix}d_x \\ d_y \\ d_z\end{bmatrix} =& \begin{bmatrix}
cos\theta & -\sin\theta & 0 \\
 sin\theta & \cos\theta & 0 \\
 0 & 0 & 1
\end{bmatrix}
\begin{bmatrix}
cos\varphi & 0 & -\sin\varphi \\
 0 & 1 & 0 \\
 sin\varphi & 0 & \cos\varphi 
\end{bmatrix}
\\
&\begin{bmatrix}
1 & 0 & 0 \\
 0 & 1 & 0 \\
 0  & 0 & 1 
\end{bmatrix} \begin{bmatrix}1 \\ 0 \\ 0\end{bmatrix} = \begin{bmatrix}\cos\theta\cos\varphi \\ \sin\theta\cos\varphi\\ \sin\varphi\end{bmatrix}
\end{aligned}
\end{equation}
It is important to note that the Euler angle transformation used here is based on a local coordinate system, and the order of matrix multiplication is the opposite of that in a global coordinate system.

\begin{figure}[!]
\begin{center}
\centerline{\includegraphics[width=\linewidth]{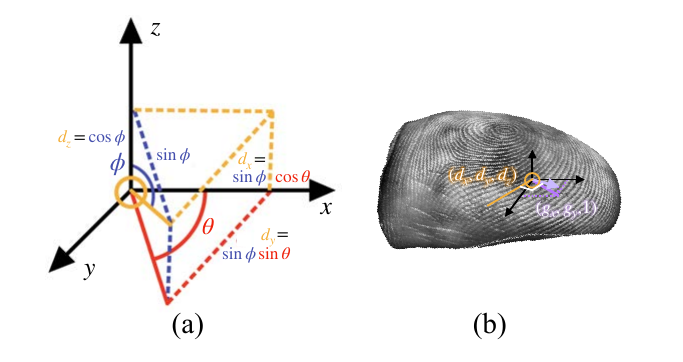}}
\vspace{-25pt} 

\end{center}
\caption{Illustration of calculating 3D minutiae. (a) represents the orientation of 2D/3D minutiae in a spherical coordinate. (b) shows a 3D minutia and the surface normal $\vec{n}$ perpendicular to it.}
\label{fig:axis}
\vspace{-12pt} 

\end{figure}

Fig. \ref{fig:axis} illustrates the calculation process of the 3D minutia, where $\phi=\frac{\pi}{2}-\varphi$. To make the scales of minutiae direction and position comparable, we multiply the minutiae direction by a constant $n$, where $n$ is set to 25.

\begin{figure*}[htb]
\begin{center}
\centerline{\includegraphics[width=\linewidth]{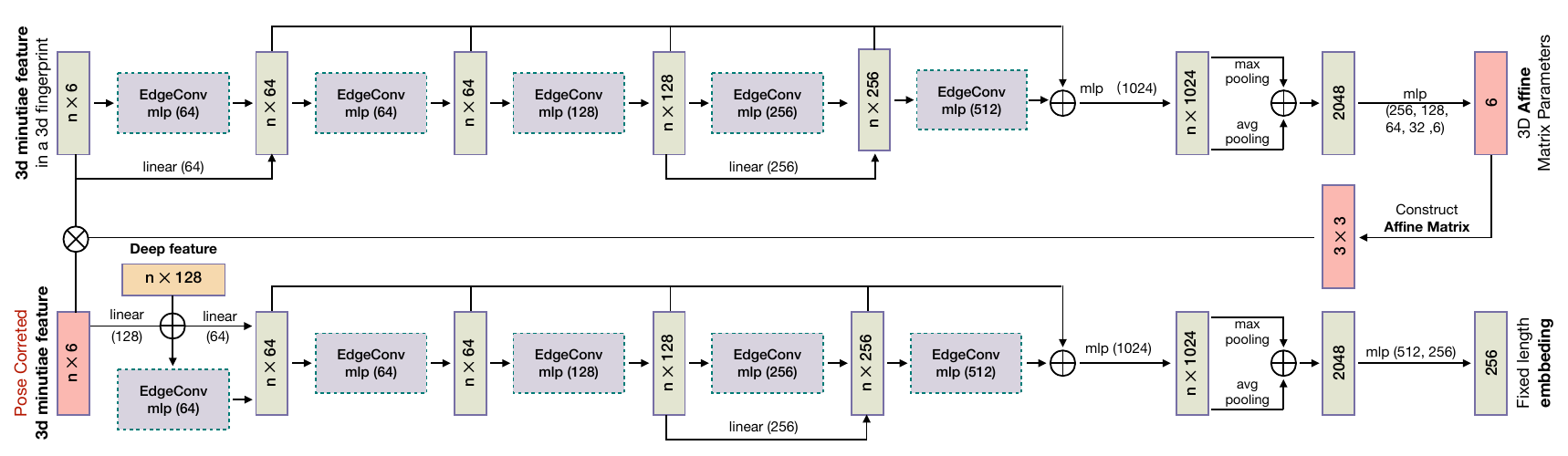}}
\vspace{-20pt} 

\end{center}
\caption{The architecture of our 3D graph matching network. The specific architecture of EdgeConv can be found in \cite{wang2019dynamic}.}
\label{fig:dgcnn}
\vspace{-10pt} 

\end{figure*}

\textbf{Graph Matching Network.} \label{ssection:3d_graph}
Our graph matching network is inspired by \cite{su2023gnn}. We use its Topological Relation Reasoning Module (TRM) module, and simplify this module just like the classification model of DGCNN \cite{wang2019dynamic}. Through EdgeConv, the network adopts a convolution-like approach, merging the 3D minutiae around each minutia in the set. 


The detailed architecture of our network is shown in Fig. \ref{fig:dgcnn}. Our network consists of two parts: the spatial transformation network and the graph embedding network. The spatial transformation network predicts and corrects the posture of 3D minutiae in 3D space. The graph embedding network extracts the 3D minutiae with corrected posture and deep features into 256-dimensional fixed-length features for matching. We project the 3D minutiae to 128-dim and sum with the deep features into a 128-dimensional input for the graph embedding network. 


We adopt a spatial transformation network, since the frequency of contactless fingerprints has been unified during preprocessing. We only need to correct the translation and rotation of 3D minutiae. After passing through five MLP layers, we obtain a 6-dimensional feature. The first three dimensions correspond to the translation in three-dimensional space $\mathbf{t}=(x_t,y_t,z_t)$, and the last three dimensions correspond to the rotation in the form of Euler angles in the $xyz$ order in three-dimensional space $(\theta_t,\psi_t,\phi_t)$. The six dimensions are then used to calculate the three-dimensional rotation matrix $\mathbf{R}$ using $\mathbf{R} = \mathbf{R}_z^{\phi_t}\mathbf{R}_y^{\psi_t}\mathbf{R}_x^{\theta_t}
$, in which $\mathbf{R}_x^{\theta_t}$, $\mathbf{R}_y^{\psi_t}$, and $\mathbf{R}_z^{\phi_t}$ represent the rotation of 3D minutiae along the $x$, $y$, and $z$ axes, respectively.

Assuming there are $N$ 3D minutiae in a contactless fingerprint image, and the positions of those 3D minutiae are represented as $\mathbf{P}=(\mathbf{p}_1^{\rm T},\dots,\mathbf{p}_N^{\rm T}) \in \mathbb{R}^{3\times N}$, and their orientations are represented as $\mathbf{O}=(\mathbf{o}_1^{\rm T},\dots,\mathbf{o}_N^{\rm T}) \in \mathbb{R}^{3\times N}$. After contactless fingerprint pose correction, a set of 3D minutiae $\mathbf{M}_t$ can be calculated according to Equation \ref{equ:posture correction}.
\begin{equation}
\mathbf{M}_t = (\mathbf{R}\mathbf{P}+\mathbf{t},\mathbf{R}\mathbf{O}+\mathbf{t})^{\rm T} \in \mathbb{R}^{N\times 6}
\label{equ:posture correction}
\end{equation}

Then, $\mathbf{M}_t$ and 3D feature $\mathbf{F}$ are fused to 128-dimensional, and are sent to the graph embedding network. After passing through two MLP layers, we obtain a 256-dimensional feature for the graph embedding network. Then, we send fixed length feature to calculate loss. The loss function is the same as the loss function in \cite{su2023gnn}, which is a triplet loss \cite{schroff2015facenet}:
\begin{equation}
Loss(a,p,n) = \max(\Vert( \boldsymbol{a},\boldsymbol{p})\Vert_2-\Vert(\boldsymbol{a},\boldsymbol{n})\Vert_2+\gamma,0).
\label{equ:triplet loss}
\end{equation}
The constant $\gamma$ is the margin. We input a batch containing pairs of fingerprints $\boldsymbol{a}$ and $\boldsymbol{p}$ matched with each other. Within batch  $\boldsymbol{a}$, we identify a batch of non-matching fingerprints  $\boldsymbol{n}$ for each fingerprint in $\boldsymbol{a}$ by selecting the closest one that is not a match. For choosing the closest fingerprint, we use L2 distance as the metric, as our loss function also computes distance using L2 distance.

Similar to the method of Cui et al. \cite{cui2023monocular}, we use our depth prediction results to unwarp the fingerprint images and perform fingerprint matching using VeriFinger. We then fuse the matching score with the graph matching similarity score to achieve state-of-the-art matching performance. If the VeriFinger matching score is denoted as $s_1$ and the graph matching score as $s_2$, the fused score is calculated as $s_1 / 1000 + s_2 / 40$.

\section{Experiment Results}
\subsection{Datasets}

The primary contactless datasets we use are UWA Benchmark 3D/2D Fingerprint Database \cite{zhou2014benchmark}, PolyU 3D+ Database \cite{lin2017tetrahedron}, CFPose Fingerprint Database \cite{tan2020towards} and ZJU Databases \cite{grosz2022contact}.

\textbf{UWA Database} \cite{zhou2014benchmark} contains 8,958 contactless fingerprints with pose variations. The ground-truth depth maps are reconstructed using the shape-from-silhouette algorithm in \cite{cui2023monocular}. The ground-truth orientation, period are obtained by VeriFinger 13.1 \cite{VeriFinger}. The last 2,988 fingerprints are used to train and validate the 3D feature extraction and graph embedding network, and the first 5,970 fingerprints are used for feature extraction and matching experiments. Since our 3D feature network needs to extract fingerprint minutiae, and the minutiae predicted by VeriFinger are often not accurate enough, we randomly selected 276 images from the training set and manually adjusted the minutiae predicted by VeriFinger as ground truth. These adjusted minutiae are used to fine-tune the minutiae prediction part of the 3D feature network.

\textbf{PolyU 3D+ Database} \cite{lin2017tetrahedron} We utilized the 2,016 ground truth depth maps and corresponding images from the first session for depth estimation experiments. This dataset, distinct from the UWA database, provides ground truth depth, better illustrating the effectiveness of our method.

\textbf{CFPose} \cite{tan2020towards} contains 1,400 contactless fingerprints with random pose variations. The ground-truth depth maps are reconstructed using the ellipsoid model algorithm in \cite{tan2020towards}. The last 200 fingerprints are used to train and validate the 3D feature extraction and graph embedding network, and the first 1,200 are used for matching experiments. The 40 ground truth minutiae manually annotated by Tan and Kumar \cite{tan2020towards} are also used to fine-tune the minutiae component of the 3D feature network.

\textbf{ZJU Database} \cite{grosz2022contact} contains 9,888 contactless fingerprints. We manually annotated the first 912 images for training the segmentation network.

\textbf{Other Database} 39,461 ids and 602,577 images in NIST301 \cite{sd301}, NIST302 \cite{sd302}, CASIA-Fingerprint v5 \cite{casiav5}, PrinsGAN \cite{engelsma2022printsgan}, and SpoofGAN \cite{grosz2022spoofgan} are also used for pretraining the matching network but not for evaluation.

\subsection{Implementation details}
Our feature extraction network is trained on 2,988 fingerprints from the UWA dataset. After preprocessing the original images and computing orientation fields, periodic maps, surface depth gradients, and minutiae, we extract two 512×512 patches from each original image, mask, and label. In total, there are 24,014 patches. The network is initially trained on the UWA dataset with minutiae extracted by VeriFinger and then finetuned with the addition 2,670 patches of the UWA and CFPose dataset with ground truth minutiae only for the Minutiae Decoder and Local MinNet. We use the Adam optimizer with a learning rate of 0.0001, $\beta_1$ of 0.9, $\beta_2$ of 0.999, and a decay rate of 0. To locate minutiae, we set $\tau_{init}=0.4$ and $\tau_{area}=45$ for minutiae extraction experiment and $\tau_{init}=0.2$ and $\tau_{area}=30$ for matching.

Our 3D graph matching network is trained using the Adam optimizer with optimizer parameters: learning rate= 0.001, $\beta_1 = 0.9$, $\beta_2 = 0.999$, weight decay= 5e-4. For batch training, we pad the number of 3D feature to 200 on each fingerprint with zeros during pretraining and finetuning. The batch size is 128. We train 80 epochs for pretraining and 100 epochs each network for finetuning. During pre-training, we project 2D fingerprint minutiae onto a 3D sphere to obtain a 6-dimensional input and use DMD \cite{pan2024latent} to get 2D fixed length deep feature around minutiae. 


Our experiments are conducted on two Nvidia 4090 GPUs and an Intel Xeon Gold 6133 CPU @ 2.50GHz.

\subsection{Evaluation of Feature Network}
The proposed feature extraction network is evaluated on the testing set from UWA and PolyU 3D+ databases for depth estimation and CFPose database for minutiae extraction. We do not evaluated orientation and period because the orientation and period are extracted by VeriFinger and may not be accurate. The depth in the UWA database, while not ground truth, is more accurate than monocular estimation due to multi-view reconstruction. Additionally, datasets with ground truth depth are relatively scarce. For the depth estimation, evaluations are conducted from gradient accuracy and depth accuracy. For minutiae extraction, we follow the evaluation protocol in \cite{tan2020towards} for testing.

Table \ref{table:error} shows the error of gradient and reconstructed depth. The depth is reconstructed using the estimated gradient by the reconstruction method in \cite{cui2023monocular}. The units of errors are: gradient in mm per pixel, depth in mm. 
\begin{figure}[!]
\begin{center}
\centerline{\includegraphics[width=1.0\linewidth]{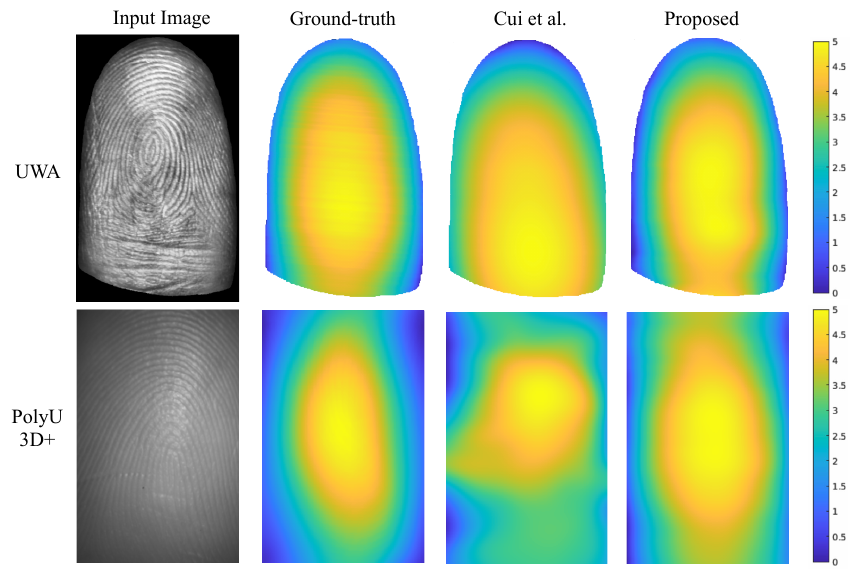}}
\vspace{-20pt} 

\end{center}
\caption{Examples of depth estimated by Cui et al.'s method \cite{cui2023monocular} and ours.}
\label{fig:depth}
\vspace{-5pt} 

\end{figure}

\begin{table}[!]
\begin{center}

\begin{tabular}{cccccc}
\toprule
\textbf{Database}&\textbf{Method}&\textbf{\tabincell{c}{Gradient\\(mm/Pixel)}}&\textbf{\tabincell{c}{Depth\\(mm)}}\\
\midrule
\multirow{2}{*}{UWA}&\tabincell{c}{Cui \emph{et al.} \cite{cui2023monocular}}&0.0321&1.4629\\

~&\tabincell{c}{Proposed}&\textbf{0.0274}&\textbf{1.1154}\\

\midrule
\multirow{2}{*}{PolyU 3D+}&\tabincell{c}{Cui \emph{et al.} 
\cite{cui2023monocular}}&0.0290&2.9577\\

~&\tabincell{c}{Proposed}&\textbf{0.0208}&\textbf{1.3455}\\

\bottomrule
\end{tabular}

\end{center}
\caption{Evaluation results of the depth network on the UWA and CFPose database}
\label{table:error}
\vspace{-10pt} 

\end{table}

Fig. \ref{fig:depth} provides an example of comparing the results of depth prediction between our method and Cui et al.'s method. Although Cui et al.'s method can predict relatively smooth depth on UWA, it struggles to capture depth variations. For instance, in the UWA example image, the depth changes sharply and shows a decreasing trend at the bottom, but Cui et al.'s method still predicts an increasing depth trend. Additionally, the PolyU 3D+ example demonstrates that Cui et al.'s method fails to generalize to data outside the training set, resulting in failed predictions.

\begin{table}[!]
\begin{center}
{\setlength{\tabcolsep}{3pt} 
 
\begin{threeparttable}
\begin{tabular}{ccccc}
\toprule
\textbf{Database}&\textbf{Method}&\textbf{Precision}&\textbf{Recall}&\textbf{F1-score}\\
\midrule
\multirow{3}{*}{CFPose}&\tabincell{c}{VeriFinger}&63.8\%&\textbf{79.1}\%&70.1\%\\
~&\tabincell{c}{Tan and Kumar \cite{tan2020towards}}$^\star	$&79.0\%&76.4\%&\textbf{77.2}\%\\

~&\tabincell{c}{Proposed}&\textbf{80.6}\%&74.1\%&76.7\%\\
\midrule
\multirow{3}{*}{UWA}&\tabincell{c}{VeriFinger}&51.7\%&\textbf{88.1}\%&64.7\%\\
~&\tabincell{c}{Tan and Kumar \cite{tan2020towards}}$^\star	$&69.1\%&74.1\%&71.0\%\\
~&\tabincell{c}{Proposed}&\textbf{74.4}\%&72.5\%&\textbf{73.0}\%\\

\bottomrule

\end{tabular}
\end{threeparttable}
}
\begin{tablenotes}
\item {$\star$}: \cite{tan2020towards} is intentionally designed for minutiae extraction, while our's is a multi-feature network. 
\end{tablenotes}
\vspace{-5pt} 

\end{center}

\caption{Evaluation results of minutiae extraction on the UWA and CFPose database}
\vspace{-15pt} 

\label{table:minu_cfpose}

\end{table}

We freeze the other parameters and only use the annotated ground truth minutiae to train the Minutiae Decoder and Local MinNet, and tested our minutiae prediction results on the ground truth dataset provided by Tan and Kumar \cite{tan2020towards}. A minutiae prediction is considered correct if its position deviated from the ground truth by 16 pixels. As Table. \ref{table:minu_cfpose} shown, the experimental results show that our method achieves a significantly higher accuracy in minutiae prediction compared to VeriFinger, demonstrating that our approach can reliably extract contactless fingerprint minutiae. By incorporating more training data from the UWA dataset, our method enables a single multi-task network to achieve higher minutiae prediction accuracy on the UWA dataset compared to ContactlessNet proposed by Tan and Kumar \cite{tan2020towards}. 

Evaluation of these features is just a references here to exam whether the network has produced reasonable features for the matching algorithm.

\subsection{Evaluation of Fingerprint Matching}

Matching experiments are performed on UWA \cite{zhou2014benchmark} and CFPose \cite{tan2020towards}. Contactless fingerprint matchings are performed, as the proposed method aims at utilizing 3D features to improve contactless fingerprint matching performances, which is in contrast to many previous methods that aim at contactless-contact matching. These CL2C methods also do not solve the pose variation issue in CL2CL matching. We compute similarity of two fingerprints through the product of corresponding embedding, which is
\begin{equation}
    s(F_i,F_j) = <M_{F_i},M_{F_j}>
\end{equation}

where $F_i$ and $F_j$ denote two fingerprints, $M_{F_i}$ and $M_{F_j}$ are the fingerprint feature embedding extracted by our graph matching network.

\begin{figure}[htb]

\centerline{\includegraphics[width=1.0\linewidth]{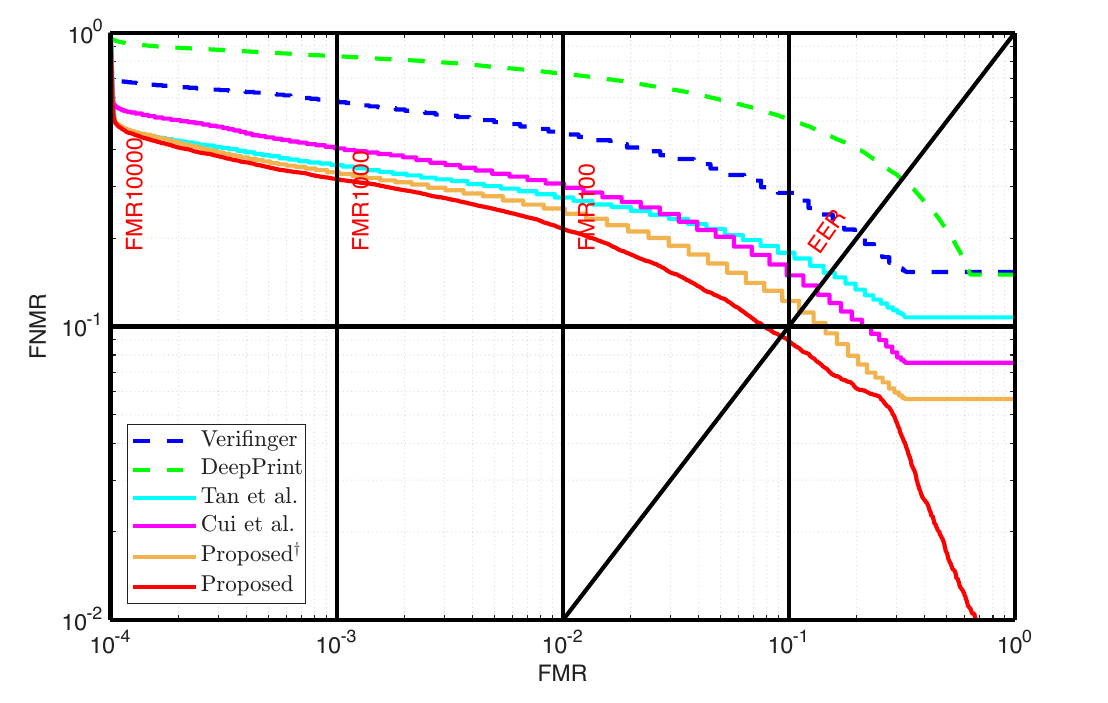}} 

\caption{DET curves on UWA database} 

\label{fig:det_curves}  
\vspace{-5pt} 

\end{figure}

We compare our matching results with the current SOTA contactless-concactless fingerprint matching algorithms, which are shown in Table \ref{table:matching}. We conduct experiments on UWA according to Cui et al.'s \cite{cui2023monocular} protocol and on CFPose and ZJU according to Dong and Kumar's \cite{dong2023syn} protocol.  The DET curve on UWA database is shown in Fig. \ref{fig:det_curves}.

We compare our method with classical fingerprint recognition approaches \cite{VeriFinger}\cite{rohwedder2023benchmarking}  and the recent deep learning methods \cite{cui2023monocular}\cite{tan2020towards}. Using the same pretrained and finetuning data, we train on an open-source DeepPrint \cite{rohwedder2023benchmarking} and perform matching experiments. We reproduce \cite{cui2023monocular}\cite{tan2020towards} for comparison. Due to the method \cite{cui2023monocular} failing to predict reliable depth on the ZJU and CFPose databases, and also no pose variation for method \cite{tan2020towards} to correct on ZJU, we do not reproduce them.

\begin{table}[!t]
\begin{center}

\begin{tabular}{lccc}
\toprule

\multirow{2}{*} {\textbf{Method}}                & \multicolumn{3}{c}{\textbf{EER}}                                                           \\ \cmidrule(lr){2-4} 
~                                        & \multicolumn{1}{c}{\textbf{CFPose}\cite{tan2020towards}}     & \multicolumn{1}{c}{\textbf{UWA}\cite{zhou2014benchmark}} & \multicolumn{1}{c}{\textbf{ZJU}\cite{grosz2022contact}} \\ \midrule

DeepPrint \cite{rohwedder2023benchmarking}             & 5.72\%           & 28.20\%           &     6.43\%                    \\
VeriFinger \cite{VeriFinger}             & 5.44\%           & 20.32\%          & 0.53\%                       \\
Tan and Kumar \cite{tan2020towards}            & 1.79\%           & 15.22\%           &       -              \\
Cui \emph{et al.} \cite{cui2023monocular}            & -              & 13.40\%           & -                       \\
Dong and Kumar \cite{dong2023syn}            & 1.12\%              & 13.84\%$^*$           & 0.48\%                       \\
Proposed$^\dag$               & \textbf{0.91\%}   & \textbf{11.16\%}   & \textbf{0.44\%}  \\
Proposed                & \textbf{0.89\%}   & \textbf{9.23\%}   & \textbf{0.37\%}                     
\\
\bottomrule
\end{tabular}
\begin{tablenotes}
\item {$*$}: A different test protocol with ours.

\item {$\dag$}: Only unwarp by our predicted depth.
\end{tablenotes}
\end{center}
\vspace{-5pt} 

\caption{Matching results on three databases}
\vspace{-5pt} 

\label{table:matching}
\end{table}

First, following the approach of Cui et al. \cite{cui2023monocular}, we unwarp the fingerprints based on the depth predicted by our 3D feature net and perform matching using VeriFinger 13.1. At this stage, we have already achieved state-of-the-art EER on all three datasets. To demonstrate the integration of 3D features into fingerprint matching, we fuse our graph matching scores with those from VeriFinger, resulting in a significant boost in matching performance. Notably, when tested under the protocol of another CL2CL method \cite{dong2023syn}, our
EER is 12.07\% compare to there 13.84\%, further proving our method’s effectiveness. Our method better identifies fingerprints with large pose variations especially on the UWA dataset with significant pose variations. This improvement shows our method effectively models large pose changes in contactless fingerprints, adding more matching information. 

\begin{figure}[!]
\begin{center}
\centerline{\includegraphics[width=1\linewidth]{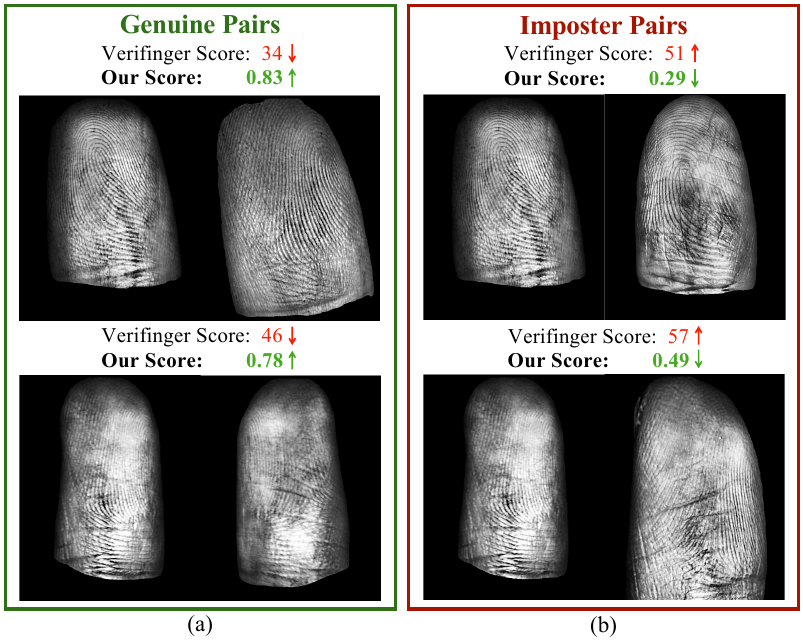}}
\vspace{-20pt} 

\end{center}
\caption{Two pairs of matching result. (a) Genuine matching fingerprints have viewpoint difference. (b) Impostor matching fingerprints have viewpoint difference.}
\label{fig:matching_example}
\vspace{-15pt} 
\end{figure}

Fig. \ref{fig:matching_example} shows genuine and impostor examples of our matching scores compared to VeriFinger's. Our graph matching method can better capture the internal information between genuine contactless fingerprints with exaggerated poses. So, it can get a high score on genuine pairs and low score on imposter pairs with similar textures, while VeriFinger is more likely to be misled by fake matching fingerprints with similar textures.

\section{Conclusion}
In this paper, we propose a novel contactless fingerprint recognition method that uses 3D graph matching to implement a 3D matching algorithm for contactless fingerprints. In contrast, previous methods only match fingerprints in 2D space and neglect the 3D information during matching. Experiment results on three databases prove that our method reaches state-of-the-art matching performances, and can successfully match contactless fingerprints across different poses.










{\small
\bibliographystyle{ieee}
\bibliography{egbib}
}

\clearpage
\appendix
\section*{Supplementary Material}

\section{Implementation Details}

\subsection{Preprocessing}


\textbf{Contrast}. We initially selected the first phalanx as foreground region. The maximum convex hull of these regions is identified as a mask, followed by the application of contrast-limited adaptive histogram (CLAHE) \cite{zuiderveld1994contrast} to normalize the contrast of the segmented images while minimizing the effect of noise. The grid size of CLAHE is empirically set to 60 pixels.

\textbf{Pose}. The finger's shape allows us to directly obtain the centerline of the finger through the contour in the preprocessing stage. Rotating the finger's centerline to a vertical position eliminates rotations along the $z$-axis, providing a certain level of normalization.

\textbf{Scale}. The scale of contactless fingerprint capture varies significantly, leading to substantial variations in the frequency of captured fingerprint ridge lines. It is necessary to normalize ridge frequency, which could affect fingerprint recognition. Thus, the ridge frequency in the central region of the fingerprint is balanced to 10 pixels across fingerprint images, avoiding affecting the subsequential feature extraction and matching steps. The average ridge period within the central area is computed using a traditional algorithm \cite{hong1998fingerprint}. However, due to the low quality of contactless fingerprints, the extracted period can sometimes be too large or too small. When the scaling factor calculated based on the frequency is greater than 1.3 or less than 1, we consider the frequency extraction to have failed. In this case, we compute the scaling factor based on the mask, setting it to three-quarters of the fingerprint image width. 

The final output fingerprint size is $640\times480$.

\subsection{Graph Embedding Network}

To train a graph network, we need many contactless fingerprints with 3D minutiae features. However, existing contactless fingerprint databases cannot meet the need for vast 3D ground-truth data. Therefore, we conduct a two-stage training strategy: first we pretrain our network on the 2D fingerprints, and finetune it on contactless 3D fingerprints. 

\begin{figure}[!]
\centering
\centerline{\includegraphics[width=0.6\linewidth]{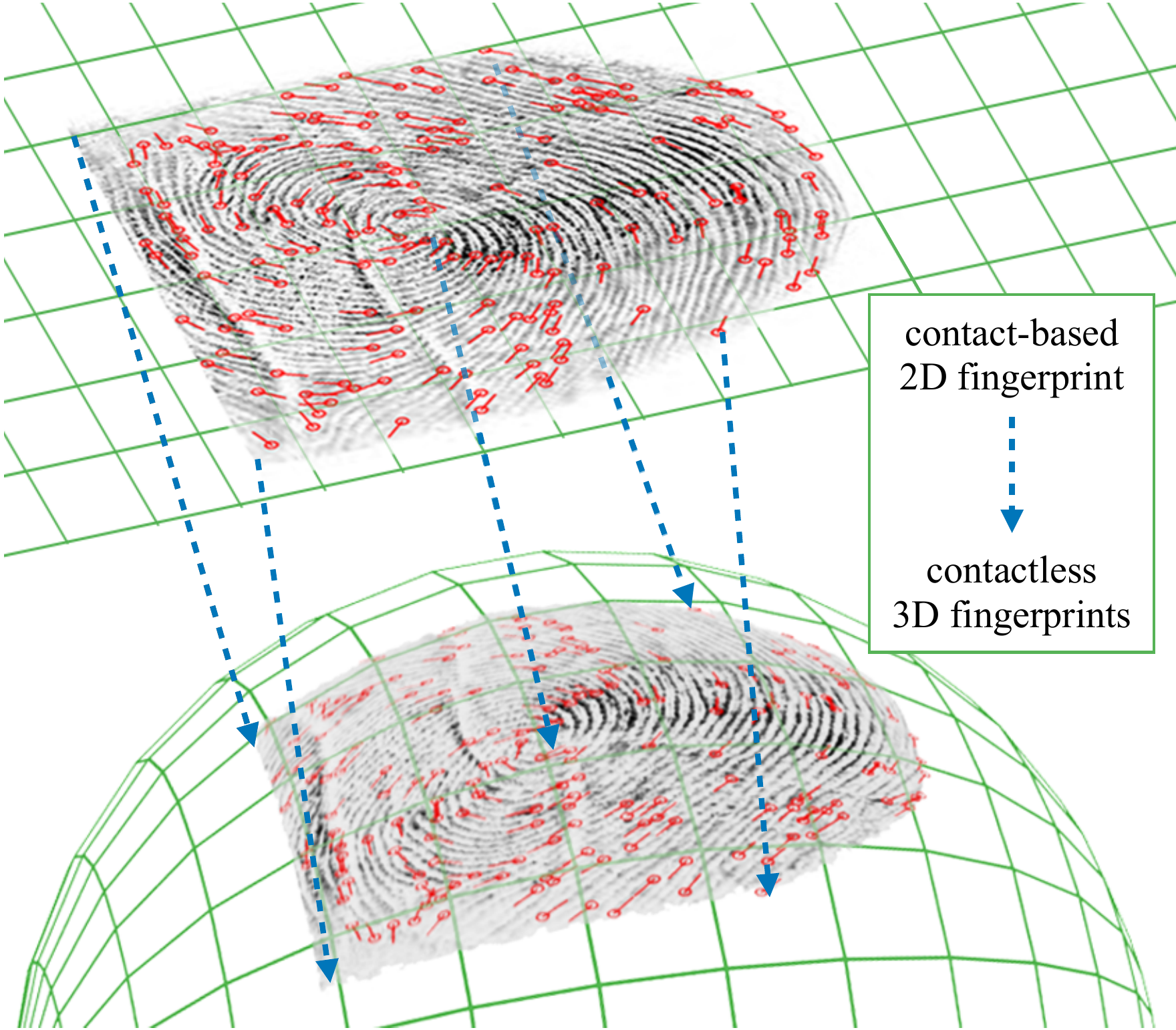}}
\caption{Illustration of generating 3D fingerprint from 2D fingerprint.}
\label{fig:3d_to_2d}
\end{figure}

In stage I, we collect a vast number of plain 2D fingerprints from NIST301 \cite{sd301}, NIST302 \cite{sd302}, CASIA-Fingerprint v5 \cite{casiav5}, PrinsGAN \cite{engelsma2022printsgan}, and SpoofGAN \cite{grosz2022spoofgan}. Then, we transfer these contact-based 2D fingerprints to 3D space to create synthetic contactless 3D fingerprints. However, given a contact-based 2D fingerprint, there are barely any methods of transferring a contact-based 2D fingerprint to a 3D fingerprint. Therefore, our research adopts a rough but efficient approximation method to project 2D fingerprint to 3D space. As Fig. \ref{fig:3d_to_2d} shows, we generally assume that the finger skin that touches the sensor is a sphere, and the contact-based 2D fingerprint is a projection of the 3D skin surface on a sphere.

\begin{figure*}[htb]
\begin{center}

\subfloat{\includegraphics[width=0.45\linewidth]{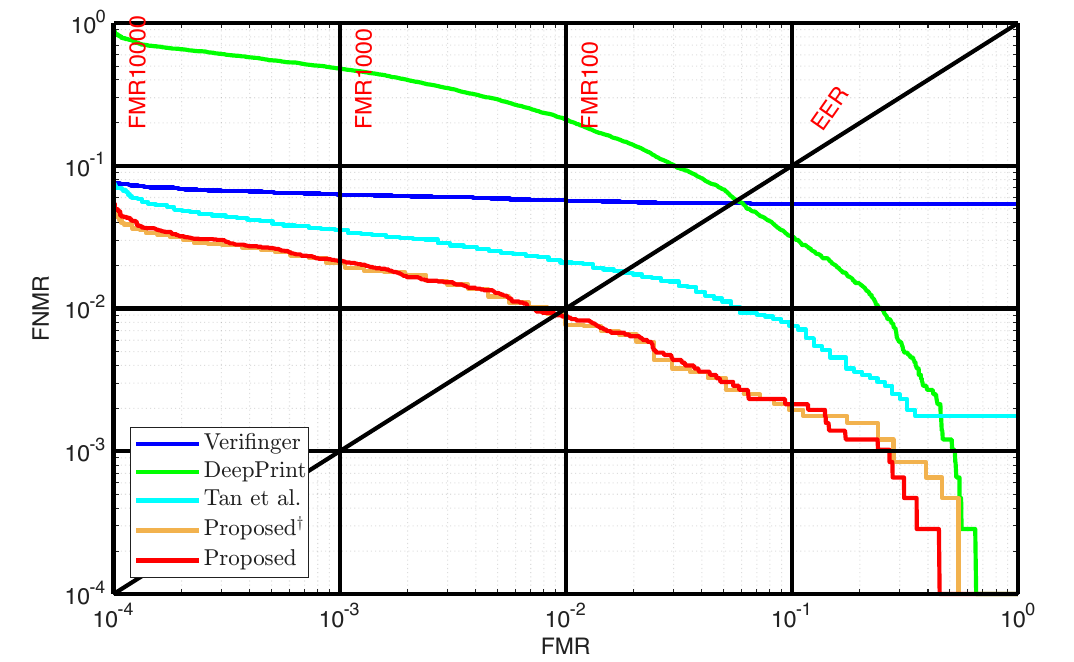}}
\subfloat{\includegraphics[width=0.45\linewidth]{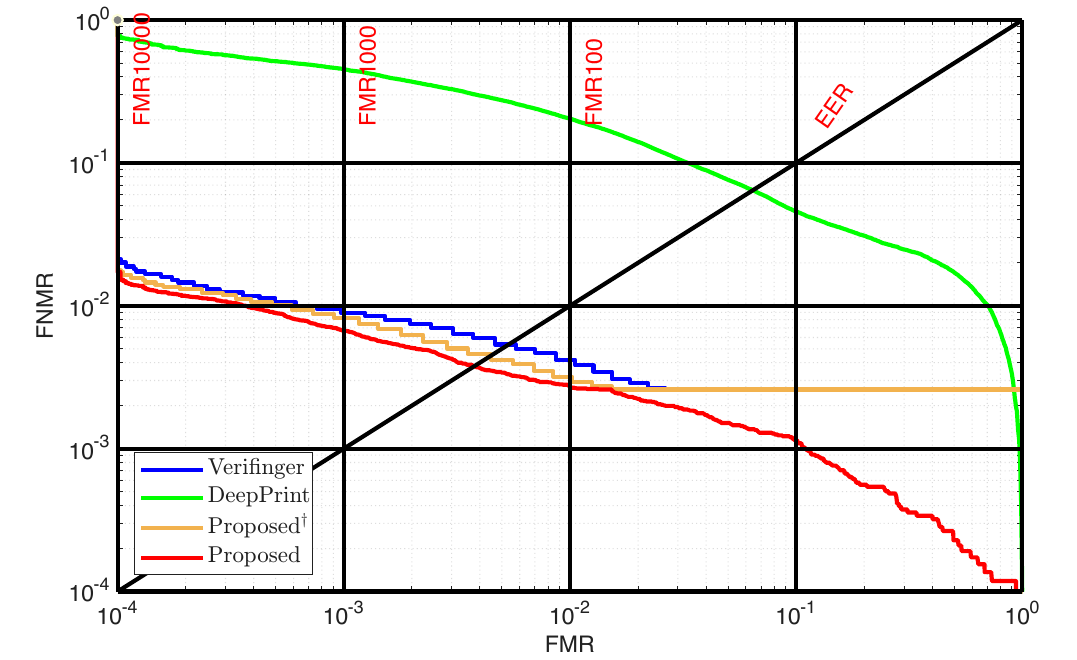}}
\end{center}

\caption{DET curves on CFPose (left) and ZJU (right) database} 

\label{fig:det_curves_supp}  

\end{figure*}

Explicitly speaking, assume the projection center is the center point of the fingerprint, the distance from a minutia to the center point is $(x,y)$, and the sphere radius is $r$. Then, its corresponding $z$-coordinate should be:
\begin{equation}
z=\sqrt{r^2-x^2-y^2}.
\label{equ:z_2d_minu1}
\end{equation}
In our implementation, we use an empirical adaptation: 
\begin{equation}
            z = \sqrt{\max_{i=1}^n(x_i^2+y_i^2) + c - x^2 -y^2}.
            \label{equ:z_2d_minu2}
\end{equation}
$c$ is a constant, which is set to 70000 here. And the 3D orientation $(d_x,d_y,d_z)$ can be easily computed since it is a spherical surface.

In stage II, the data used for finetuning is the 3D minutiae based feature points from contactless fingerprints. We extract some minutiae based feature points via our 3D feature net. Specifically, we adjust the recall threshold  $\tau_{init}$  for minutiae. $\tau_{init}$ is 0.4 when recalling minutiae and 0.2 when recalling 3D minutiae based feature points.

Due to the higher frequency of contactless fingerprints with more minor pose variations in real-life scenarios, we sample the fingerprint pairs in the same direction three times. The fingerprint pairs of front and left poses, as well as front and right poses, are sampled twice, making the network easier to train.

For the CFPose database, we adopt the last 200 fingerprints for training. We perform three-times samplings for fingerprint pairs with a yaw angle difference $\Delta{\theta}\leq10^\circ$ along the y-axis, six-times samplings for those with $10^\circ<\Delta{\theta}\leq45^\circ$, and three-times samplings for the rest, achieving data balance within the CFPose dataset as well as the entire training set. The yaw angle is estimated by the method in \cite{tan2020towards}.

We perform data augmentation on the input 3D minutiae in finetuning, including random 3D rotation (all axes range from $-15^\circ$ to $15^\circ$) the fingerprints with $\Delta{\theta}\leq10^\circ$. For the all fingerprints, we also perform random translation (all three axes range from $-100$ to $100$ pixels), and random dropping of $1/6$ of all the 3D minutiae based feature points.

\section{Supplement to the Feature Extraction Experiments}

We present minutiae extraction accuracy at 8-pixel resolution in Table \ref{table:minu_cfpose_supp}, and the results demonstrate the same conclusions as those observed at 16-pixel resolution.

\begin{table}[!]
\begin{center}
{\setlength{\tabcolsep}{3pt} 
 
\begin{threeparttable}
\begin{tabular}{ccccc}
\toprule
\textbf{Database}&\textbf{Method}&\textbf{Precision}&\textbf{Recall}&\textbf{F1-score}\\
\midrule
\multirow{3}{*}{CFPose}&\tabincell{c}{VeriFinger}&55.9\%&69.4\%&61.4\%\\
~&\tabincell{c}{Tan and Kumar \cite{tan2020towards}}&75.9\%&\textbf{73.3}\%&\textbf{74.1}\%\\

~&\tabincell{c}{Proposed}&\textbf{76.9}\%&70.8\%&73.2\%\\
\midrule
\multirow{3}{*}{UWA}&\tabincell{c}{VeriFinger}&51.7\%&\textbf{74.9}\%&54.9\%\\
~&\tabincell{c}{Tan and Kumar\cite{tan2020towards}}&65.4\%&70.1\%&67.2\%\\
~&\tabincell{c}{Proposed}&\textbf{68.7}\%&67.1\%&\textbf{67.4}\%\\

\bottomrule

\end{tabular}
\end{threeparttable}
}
\vspace{-5pt} 

\end{center}

\caption{Minutiae extraction at 8-pixel threshold results}

\label{table:minu_cfpose_supp}

\end{table}

\section{Supplement to the Matching Experiments}
We conduct experiments on UWA separately according to Cui et al.'s \cite{cui2023monocular} protocol which performs 3,000$\times$3,000 matchings. There are 9,000 pairs of genuine matchings, and the rest of the pairs are impostor matching. The missing partial matching scores are set to zero. For evaluating the matching performance on CFPose, a total of 719,400 pairs of matchings are performed. 5,400 pairs are genuine matching, and the rest are impostor matching. As segmentation has minimal impact on fingerprint matching, we still use all 9,888 fingerprints for matching experiment. Following \cite{dong2023syn}, we utilize $824\times12\times11/2 = 54,384$ genuine matches and $824\times823/2 = 339,076$ impostor matches (from the first image of each id) for matching experiments.

Fig. \ref{fig:det_curves_supp} presents the DET curves of our method on the CFPose and ZJU datasets. It shows that our new depth prediction and graph matching method significantly boosts contactless fingerprint recognition, especially when unfolding CFPose for 2D matching. This improves fingerprint matching performance and addresses previous poor generalization in fingerprint depth prediction.

Table \ref{table:matching_supp} shows the ZeroFMR of different methods. Upon combining with the DET curve we provide, it can be found that our method surpasses previous ones at various FMR levels.

\begin{table}[!t]
\begin{center}

\begin{tabular}{lccc}
\toprule

\multirow{2}{*} {\textbf{Method}}                & \multicolumn{3}{c}{\textbf{ZeroFMR}}                                                           \\ \cmidrule(lr){2-4} 
~                                        & \multicolumn{1}{c}{\textbf{CFPose}\cite{tan2020towards}}     & \multicolumn{1}{c}{\textbf{UWA}\cite{zhou2014benchmark}} & \multicolumn{1}{c}{\textbf{ZJU}\cite{grosz2022contact}} \\ \midrule

Verifinger             & 7.04\%           & 66.64\%          & 1.66\%                       \\
Tan and Kumar \cite{tan2020towards}            & 5.43\%           & 44.53\%           &       -              \\
Cui \emph{et al.} \cite{cui2023monocular}            & -              & 51.99\%           & -                       \\
Proposed$^\dag$               & \textbf{3.37\%}   & \textbf{44.94\%}   & \textbf{1.39\%}  \\
Proposed                & \textbf{3.65\%}   & \textbf{43.23\%}   & \textbf{1.16\%}                     
\\
\bottomrule
\end{tabular}
\begin{tablenotes}

\item {$\dag$}: Only unwarp by our predicted depth for 2D matching.
\end{tablenotes}
\end{center}

\caption{ZeroFMR on three databases}

\label{table:matching_supp}
\end{table}

\section{Ablation Study}

\begin{table}[!]
\begin{center}
{\setlength{\tabcolsep}{12pt} 

\begin{threeparttable}
\begin{tabular}{ccccccccc}
\toprule
\textbf{a}&\textbf{b}&\textbf{c}&\textbf{d}&\textbf{e}&\textbf{f}&\textbf{EER}\\
\midrule
\checkmark & & & & \checkmark& &  21.99\%\\
 & \checkmark& && \checkmark&  & 21.46\%\\
&  \checkmark& \checkmark & & & &  18.99\%\\
&  \checkmark& \checkmark&\checkmark &\checkmark& &  21.00\%\\
&  \checkmark& \checkmark& &\checkmark& &  \textbf{16.80\%}\\
&  \checkmark& & & & \checkmark&  \textbf{11.16\%}\\

&  \checkmark& \checkmark& &\checkmark&  \checkmark&  \textbf{9.23\%}\\

\bottomrule
\end{tabular}
\begin{tablenotes}
\item a: Minutiae from Verifinger.
\item b: Minutiae from our 3D Feature Network.
\item c: Combining 128-dim deep feature for training.
\item d: Only using 2D Minutiae
\item e: Spatial Transformation Network.
\item f: 2D Score with our unwarp method.
\end{tablenotes}
\end{threeparttable}
}

\end{center}

\caption{Ablation study of training strategies on the UWA database}

\label{table:abs_module}
\end{table}

Table \ref{table:abs_module} presents the results of our ablation experiments. This ablation study focuses on the 3D matching process. First, by replacing the minutiae extracted by Verifinger with those from our 3D Feature Network, we demonstrate that our network better captures the shape of 3D fingerprints. Next, incorporating the 128-dimensional deep features from the intermediate layer of the 3D Feature Network further enhances network performance. We also show that position correction in 3D space improves performance. Finally, our method’s ability to effectively combine with other approaches significantly boosts the performance of cross-pose contactless fingerprint recognition. At the same time, our method can also capture new matching information on the ZJU dataset, which lacks large scale pose changes. It works even when FMR is high, effectively handling difficult samples.

Moreover we re-trained the 3D Graph Matching Network using 2D minutiae and evaluated it directly on the UWA dataset for matching. The 3D minutiae results below correspond to the 3D matching outcomes in our Supplementary Material when no 2D scores were fused. To align the input formats of 2D and 3D minutiae, instead of using the traditional $(x,y,\theta)$ as input, we used $(x,y,d_x,d_y)$. The experimental result shows that 3D minutiae can significantly enhance the performance of 3D graph matching.


\end{document}